\documentclass[runningheads]{llncs}
\usepackage[T1]{fontenc}
\usepackage{graphicx}

\usepackage{hyperref}
\usepackage{color}

\usepackage{xcolor}
\usepackage{amssymb}
\usepackage{multirow}
\usepackage{color, colortbl}
\usepackage{makecell}
\usepackage{gensymb}
\usepackage{mathtools}
\usepackage{algorithm,algpseudocode}
\usepackage{bm}
\usepackage[misc]{ifsym}
\usepackage{dsfont}
\usepackage{tabularx}
\usepackage{booktabs}

\begin{document}
\title{SurvCORN: Survival Analysis with Conditional Ordinal Ranking Neural Network}
\titlerunning{Survival Analysis with Conditional Ordinal Ranking Neural Network}
%
\author{Muhammad Ridzuan\orcidID{0000-0003-0935-8466} \and Numan Saeed\orcidID{0000-0002-6326-6434} \and Fadillah Adamsyah Maani\orcidID{0000-0001-5927-7782} \and
Karthik Nandakumar\orcidID{0000-0002-6274-9725} \and Mohammad Yaqub\orcidID{0000-0001-6896-1105}} 

\authorrunning{M. Ridzuan et al.}

\institute{Mohamed Bin Zayed University of Artificial Intelligence, Abu Dhabi, UAE \\
\email{\{Muhammad.Ridzuan, Numan.Saeed, Fadillah.Maani, Karthik.Nandakumar, Mohammad.Yaqub\}@mbzuai.ac.ae}}

\maketitle              
\begin{abstract}
Survival analysis plays a crucial role in estimating the likelihood of future events for patients by modeling time-to-event data, particularly in healthcare settings where predictions about outcomes such as death and disease recurrence are essential. However, this analysis poses challenges due to the presence of censored data, where time-to-event information is missing for certain data points. Yet, censored data can offer valuable insights, provided we appropriately incorporate the censoring time during modeling. In this paper, we propose SurvCORN, a novel method utilizing conditional ordinal ranking networks to predict survival curves directly. Additionally, we introduce SurvMAE, a metric designed to evaluate the accuracy of model predictions in estimating time-to-event outcomes. Through empirical evaluation on two real-world cancer datasets, we demonstrate SurvCORN's ability to maintain accurate ordering between patient outcomes while improving individual time-to-event predictions. Our contributions extend recent advancements in ordinal regression to survival analysis, offering valuable insights into accurate prognosis in healthcare settings. Our code is available at https://github.com/BioMedIA-MBZUAI/SurvCORN.

\end{abstract}

\section{Introduction}

Survival analysis estimates the probability of a future event occurring to patients by modeling time-to-event data. In healthcare settings, common medical applications include predicting the time to death, recurrence of diseases, or re-hospitalization of patients using medical images and Electronic Health Records (EHRs). Survival analysis is a challenging problem due to the presence of censored data; for certain data points, the time-to-event information is missing due to various reasons such as patients discontinuing follow-up visits, relocating, or withdrawing from a study \cite{sparr1993returns}. However, censored data can be useful if we utilize the censoring time during modeling because it entails a time until which we are sure that the event did not happen, e.g., the patient did not die. 

A commonly reported metric in survival analysis is the concordant index, which is used to evaluate the pairwise concordance of survival times \cite{Harrell1982,Harrell1984,Harrell1996}. However, it does not provide a simple, interpretable assessment of the actual time-to-event predictions. Patients and clinicians alike are likely to benefit from saying a model has a prediction error of $X$ number of days than simply saying it has a concordant index of $Y$.

Deep survival methods can be broadly categorized into continuous-time and discrete-time methods, where the discrete methods approximate the continuous models by discretizing the continuous survival time scale. Continuous methods include DeepSurv \cite{deepsurv}, built upon the semi-parametric Cox Proportional Hazard (CoxPH) \cite{coxph,deepsurv} model which assumes the ratio of the hazard functions for any two individuals is constant over time.
Discrete methods include Nnet-Survival \cite{nnet} and DeepHit \cite{deephit}. 
DeepHit \cite{deephit} parameterizes the event-time probability mass function, while Nnet-Survival parameterizes the hazard rate using a Bernoulli function.
Our proposed method falls under the discrete category.

A pivotal tool in survival analysis is the survival curve, offering a graphical representation of the fraction of patients surviving after a specific event. The survival curve is a decreasing function that allows healthcare professionals to assess the probability of survival over time and compare survival experiences between different patient groups or treatment modalities. 
Recently, Shi et al. \cite{corn2023} introduced a rank-consistent ordinal regression for neural networks based on conditional probabilities.  They provide strong theoretical guarantees for rank-monotonicity, where the rank of an object changes monotonically with its predicted probability of belonging to a higher- or lower-ranked category. 

We observe this property to be desired for survival analysis, where the predicted probability of a patient's survival decreases over time in the absence of an intervention. We thus propose an extension to this method that accounts for both censored and uncensored cases and directly predicts the survival curve using a conditional probability interpretation of the network output.
Our contributions are two-fold:
\begin{itemize}
    \item We introduce \textit{SurvCORN}, a \textit{Surv}ival analysis method using \textit{C}onditional \textit{O}rdinal \textit{R}anking \textit{N}etwork, that directly predicts patients' survival curves using conditional probabilities 
    \item We propose \textit{SurvMAE}, a metric that evaluates the quality of a model's survival predictions based on how far they are from the actual recorded time-to-events
\end{itemize}
We show empirically that SurvCORN is able to maintain a correct ordering of patient outcomes while improving upon individual time-to-event predictions. 

\section{Method}
We aim to develop a deep neural network that directly predicts a monotonically decreasing function from $1 \rightarrow 0$ to represent a patient’s survival probability over time. We first divide the time axis into $K$ number of discrete intervals, or \textit{time bins}, then use logistic regression to predict the patient's survival probability beyond each time bin. The $K$-th time bin represents a final all-encompassing time frame beyond the maximum time in the training set ($T_{max}, \infty$), i.e., right-censored at $T_{max}$. 

Given a training dataset $D = \{X^{i}, \delta^{i}, T^{i}\}^N_{i=1}$, where $N$ is the number of patients, $X^{i}$ is the patient features (e.g. MRI, CT scans, EHR), $\delta^{i}$ is the event indicator (with $\delta^{i} = 1$ indicating event occurrence for uncensored patients, while $\delta^{i} = 0$ indicating event unobserved for censored patients), and $T^{i}$ is the actual time-bin index of the discretized time-to-event. 
The objective is to train a network $f_{\theta}(X) \rightarrow \boldsymbol{z} \in \mathbb{R}^{d_{K-1}}$ to predict a patient's survival probability beyond each time bin $t_k$, denoted as $S(T>t|X) = P\left(T > t \mid X \right)$.
The output of the network is of size $K-1$, not $K$, because a patient who survives beyond $t_{K-1}$ is assumed to experience the event at $t_K$.
In other words, a high probability of surviving beyond $t_{K-1}$ means that the patient will likely experience the event at $t_K$.

\subsection{Label encodings}
Unique to survival analysis is the presence of censoring. Here, we present a separate label encoding for uncensored versus censored patients that is crucial for the minimization of the log-likelihood.

\begin{figure}
    \centering
    \includegraphics[scale=0.5]{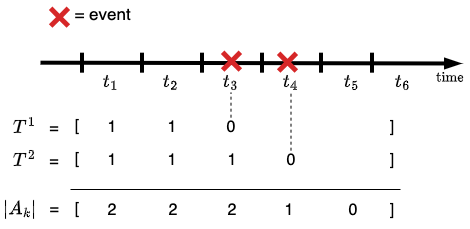}
    \caption{An illustration of the uncensored encodings of a batch of two patients who experience an event at $t_3$ and $t_4$, respectively, from a total of $K=6$ time bins. Each entry represents $T^i_k = \mathds{1}\{T^i>t_k\}$. $|A_k|$ is the size of the conditional training subsets for each time bin in the batch.}
    \label{fig:uncensored}
\end{figure}

\textbf{Uncensored encoding ($\delta^{i} = 1$).} For uncensored patients, the time bin index $T^i$ is transformed into an ordered vector representation of $K-1$ binary labels $\{T^i_{1}, T^i_{2}, …, T^i_{K-1}\}$, where $T^i_{k} = \mathds{1}\{T^i>t_k\} \in \{0,1\}$ denotes whether $T^i$ exceeds $t_k$, i.e. whether the patient $i$ survives beyond the time $t_k$, and $t_1 \prec t_2 \prec … \prec t_{K-1}$. Figure \ref{fig:uncensored} illustrates an example of $K=6$ time bins and two patient encodings whose events are recorded at $t_3$ and $t_4$, respectively.

\begin{figure}
    \centering
    \includegraphics[scale=0.5]{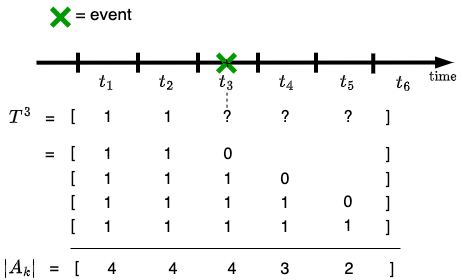}
    \caption{An illustration of the expanded censored encodings of a batch of one patient whose event information is unknown at $t_3$ onwards, from a total of $K=6$ time bins. Each entry represents $T^i_k = \mathds{1}\{T^i>t_k\}$. $|A_k|$ is the size of the conditional training subsets for each time bin in the batch. Here, the patient's survival probability beyond a time bin $t$ is $\sum{A_k}/|A_k|$, i.e., 100\% (4/4) for $t_1$ and $t_2$, indicating certainty in the patient's survival status, followed by 75\% (3/4) for $t_3$, 67\% (2/3) for $t_4$, and 50\% (1/2) for $t_5$.}
    \label{fig:censored}
\end{figure}

\textbf{Censored encoding ($\delta^{i} = 0$).} For censored patients, only the lower bound of the patients' survival times are known, i.e., that they are alive or did not experience an event at least up until the recorded time $t_k$. In this case, the actual time-to-events are unknown, and the patients have a chance of experiencing an event at any future time bin. Consider a third patient who is censored at $t_3$. Assuming independent censoring, where the patient is equally likely to experience an event as others with similar covariates,
the patient's encoding can be expanded to account for all possibilities of the event happening from the time of censoring to the final time bin (Figure \ref{fig:censored}). In presenting the model with all possible combinations of time-to-events, the model also implicitly learns that a patient's survival probability decreases over time because the ratio of survival to event occurrence also decreases.

\subsection{Network output}
The output of the network is a sequence of conditional probabilities using conditional training subsets $A_k$ at each time bin where the sigmoid output of each neuron is interpreted as:
\begin{equation}
f_{k}(X^i) = \sigma(z^i) = P(T^i>t_{k}|T^i>t_{k-1},X^i)
\end{equation}

with nested events $\{T^i > t_k\} \subseteq \{T^i > t_{k-1}\}$, where $f_k$ is the output of the $k$-th neuron, corresponding to the $k$-th time bin.

The survival output at each time bin can be computed by applying the chain rule for probabilities to the output neurons:

\begin{equation}
S(T^i>t_k|X^i) = \prod_{j=1}^k f_j(X^i)    
\end{equation}

Since each probability lies between 0 and 1, as the time bin index increases, this cumulative product guarantees a monotonically decreasing function that is desired for survival curves. To obtain the time-bin index of the prediction, we calculate 

\begin{equation}
\hat{T^i} = 1 + \sum_{k=1}^{K-1} \mathds{1}\{f_k(X^i)>0.5\}
\end{equation}

corresponding to the median of the survival curve \cite{Reid1981}. To maintain a conditional probability interpretation, we construct the conditional training subsets as proposed by \cite{corn2023} with sizes $|A_k|$ equal to the number of samples in the batch that satisfies $\{T^i>t_{k}\}$, yielding $|A_1| \geq |A_2| \geq ... \geq |A_{K-1}|$ (see Figures \ref{fig:uncensored} and \ref{fig:censored} for illustration). 

\subsection{SurvCORN loss}
The SurvCORN loss consists of two components: a log-likelihood and a ranking loss. Following the above, the log-likelihood is

\begin{equation}
loss_{LL} = - \frac{1}{\sum_{k=1}^{K-1}|A_k|} \sum_{k=1}^{K-1} \sum_{i=1}^{|A_k|} [ \mathds{1}\{T^i > t_k\} \cdot log(f_k(X^i)) + \mathds{1}\{T^i \leq t_k\} \cdot log(1-f_k(X^i))]
\end{equation}

The ranking loss directly optimizes the concordant index by penalizing incorrect ordering of pairs and is adapted from \cite{deephit}:

\begin{equation}
\label{eq:rank_loss}
loss_{rank} = \sum_{i,j} \delta^i \, \mathds{1}\{T^i < T^j\} \exp\left(\frac{\hat{S}(T^i \mid X^i) - \hat{S}(T^j \mid X^j)}{\alpha}\right)\\
\end{equation}

where $\alpha$ is a hyperparameter set to 0.1 following \cite{deephit}. The final SurvCORN loss is a summation of the two:

\begin{equation}
\mathcal{L}_{final} = loss_{LL} + loss_{rank}
\end{equation}

The log-likelihood drives the model to learn the correct survival times of each patient (i.e., \textit{intra-patient} predictions), while the ranking loss encourages the model to learn the correct ordering between patient survival times (i.e. \textit{inter-patient} predictions). 

\section{Evaluation Metric}
We report the time-dependent concordant index (C-index) of \cite{antolini2005} and introduce a new metric called SurvMAE.

The C-index is a typical metric reported in survival analysis \cite{Harrell1982,Harrell1984,Harrell1996}. It compares the ordering of every pair of patients and quantifies the proportion of pairs in the dataset whose predicted survival times are concordant with the actual survival times. A pair of individuals is considered concordant if the individual with the longer predicted survival time (or conversely, the lower predicted risk) also has the longer observed survival time. 
As Harrel's C-index \cite{Harrell1982,Harrell1984,Harrell1996} was originally derived for the proportional hazards framework, Antolini et al. \cite{antolini2005} extended it to the non-proportional cases by introducing a time-dependent variation of the C-index which we employ in this paper. 
The C-index accounts for the \textit{relative} ordering of patients, but does not take into consideration the actual predicted times of a survival model.

To this end, we propose survival mean absolute error (\textbf{SurvMAE}), a simple metric built upon MAE that accounts for different censoring mechanisms. Specifically, for censored patients whose actual time-to-events are unknown, the MAE is calculated as the average of the predicted times to the left of (i.e., less than) or equal to the actual times. For uncensored patients, the MAE is calculated normally for predictions to the left or right of the actual times. Mathematically, SurvMAE is represented by the following:

\begin{equation}
\label{eq:PLACEHOLDER} 
{SurvMAE} = \frac{1}{N_U} \sum_{i=1}^{N_U} \left [ \delta^i || \hat{T^i} - T^i ||_1 \right] + \frac{1}{N_C} \sum_{j=1}^{N_C} \left[ (1- \delta^i ) || \hat{T^j} - T^j ||_1 \cdot \mathds{1}\{\hat{T^j} \leq T^j\} \right]\\
\end{equation}

where $N_U$ is the number of uncensored patients, $N_C$ is the number of censored patients, $\delta$ is the event indicator ($0$ for censored, $1$ otherwise), and $\hat{T}$ and $T$ are the predicted and actual survival times, respectively. For continuous-time survival models, SurvMAE gives a directly interpretable prediction error in unit time (i.e., in terms of the number of days, months, etc). For discrete equidistant time bins, SurvMAE can be translated into an interpretable time-to-event metric through interpolation, where the lower the SurvMAE the better.

Reporting both C-index and SurvMAE is crucial to give a more holistic understanding of the performance of survival models. C-index provides insight regarding the correct ordering of patients, while SurvMAE evaluates the individual predicted time-to-events of the patients.

\section{Experimental Setup}

\subsection{Datasets}
We assess the prognostic ability of SurvCORN by comparing it against conventional benchmarks in analyzing two real-world cancer datasets: ChAImeleon \cite{CHAIMELEON} and HECKTOR \cite{hecktor2021a,hecktor2021b}.

\textbf{ChAImeleon} \cite{CHAIMELEON} is a lung cancer dataset comprised of CT scans and EHR from 320 different patients, with 59\% of the data being censored. The average survival time of the patients is 40.5 months with a standard deviation of 20 months. In this work, we employ a segmentation model from \cite{Hofmanninger2020} to crop the lung areas, allowing the prognostic models to focus on the region of interest.

\textbf{HECKTOR} \cite{hecktor2021a,hecktor2021b} is a multi-centric, multi-modal head-and-neck cancer dataset consisting of 224 CT and PET scans with corresponding EHRs, with 75\% of the data being censored. The average survival time of the patients is 27.8 months with a standard deviation of 24.3 months. The PET and CT scans are registered to a common origin 
to enable accurate integration of useful information from the two imaging modalities.

We preprocess the images in each dataset to standardize the prognostic model input by performing resampling, cropping, and resizing to a final input size of $112\times112\times130$. Additionally, we use equidistant binning of the survival time distribution where the number of time bins for each dataset is obtained as the square root of the number of observations.

\subsection{Implementation Details}
We run all experiments for 30 epochs using a five-fold cross-validation with a DenseNet-121 \cite{densenet} architecture, a batch size of 16, Adam \cite{adam2015} optimizer with a learning rate of $1\times10^{-3}$ and a weight decay of $1\times10^{-5}$. We maintain a constant ratio of patients who experienced each event and those who were censored in each fold. All experiments are implemented using PyTorch \cite{pytorch}. 
Since the survival methods differ primarily in their loss function and treatment of the network output, we keep the network architecture and hyperparameters fixed to ensure a fair and unbiased comparison across datasets and experiments. 

\section{Results} 
We compare SurvCORN against three deep baseline survival models (i.e. Nnet-Survival \cite{nnet}, DeepMTLR \cite{deepmtlr}, and DeepHit \cite{deephit}), all of which treat the survival times as discrete. These models have been shown to achieve competitive overall predictive performance in the survival analysis literature. We report the averages and standard deviations of the C-index and SurvMAE in Tables \ref{tab:results_lung} and \ref{tab:results_hecktor} for the ChAImeleon Lung Cancer~\cite{CHAIMELEON} and HECKTOR~\cite{hecktor2021a,hecktor2021b} datasets, respectively. 
Our method achieves competitive performance in C-index while it consistently outperforms other methods in SurvMAE.

\begin{table}[!t] 
\centering
\caption{Average concordance indices and SurvMAE ($\pm$ standard deviations) on the ChAImeleon Lung Cancer~\cite{CHAIMELEON} dataset using deep discrete-time survival models. SurvMAE (time) is calculated through a linear interpolation of SurvMAE (bin) where the survival curves hit 0.5. All experiments are run with five-fold cross-validation. 
The best scores per dataset are bolded. 
} 
    \begin{tabularx}{0.95\columnwidth}{@{\extracolsep{\fill}} lccc }
        \toprule
        & C-index $\uparrow$ & SurvMAE (bin) $\downarrow$ & SurvMAE (time) $\downarrow$ \\
        \midrule
        \textbf{Nnet-Survival~\cite{nnet}} & \textbf{0.655} $\pm$ 0.07 & 4.55 $\pm$ 0.71 & 62.7 $\pm$ 12.7 \\

        \textbf{DeepHit~\cite{deephit}} & 0.625 $\pm$ 0.06 & 4.62 $\pm$ 1.25 & 64.0 $\pm$ 13.7 \\
        

        \textbf{SurvCORN (ours)} & 0.633 $\pm$ 0.05 & \textbf{4.27} $\pm$ 0.51 & \textbf{52.9} $\pm$ 24.9 \\ 
        \bottomrule
    \end{tabularx}
    \label{tab:results_lung}
\end{table}

\begin{table}[!t] 
\centering
\caption{Average concordance indices and SurvMAE ($\pm$ standard deviations) on the HECKTOR~\cite{hecktor2021a,hecktor2021b} dataset using deep discrete-time survival models. All experiments are run with five-fold cross-validation. SurvMAE (time) is calculated through a linear interpolation of SurvMAE (bin) where the survival curves hit 0.5. The best scores per dataset are bolded. 
}
    \begin{tabularx}{0.95\columnwidth}{@{\extracolsep{\fill}} lccc }
        \toprule
        & C-index $\uparrow$ & SurvMAE (bin) $\downarrow$ & SurvMAE (time) $\downarrow$ \\
        \midrule
        \textbf{Nnet-Survival~\cite{nnet}} & 0.683 $\pm$ 0.03 & 6.20 $\pm$ 0.50 & 38.4 $\pm$ 10.0 \\

        \textbf{DeepHit~\cite{deephit}} & 0.674 $\pm$ 0.04 & 6.02 $\pm$ 0.25 & 57.3 $\pm$ 58.1 \\
        
        
        \textbf{SurvCORN (ours)} & \textbf{0.705} $\pm$ 0.04 & \textbf{5.06} $\pm$ 0.08 & \textbf{32.6} $\pm$ 2.5 \\ 
        \bottomrule
    \end{tabularx}
    \label{tab:results_hecktor}
\end{table}

\section{Discussion and Conclusion}

Compared to Nnet-Survival \cite{nnet}, DeepHit \cite{deephit} performs slightly worse, consistent with the findings in literature (e.g. \cite{kvamme}). 
For censored data, Nnet-Survival \cite{nnet} allows the model to operate without penalization beyond the recorded survival time, while SurvCORN implicitly models a decreasing survival probability beyond the recorded survival time. 
The conditional probability formulation of SurvCORN encourages the model to learn predictions that closely approximate the actual time-to-events, thus optimizing for SurvMAE. Additionally, like DeepHit \cite{deephit}, SurvCORN employs a ranking loss to promote the correct ordering of survival times, thus optimizing for C-index.


Our model, SurvCORN, not only competes effectively with existing baselines in terms of C-index predictions but also demonstrates superior performance in SurvMAE scores. This distinction is particularly noteworthy as the C-index has been deemed unreliable, uninterpretable, and clinically less useful in many works (e.g. \cite{ci_pitfall_Cook2007,ci_pitfall_Vickers2010,ci_pitfall_Hartman2023}). 
SurvMAE evaluates the accuracy of the predicted time-to-events of the patients, thus reflecting our model's precision in estimating event occurrences. This holds significance in prognosis outlook prediction as it gives clinicians timely insights to make informed decisions on treatment plans, resource allocation, and patient management strategies.

\paragraph{Limitations.}
Like other discrete survival methods, SurvCORN does not allow extrapolation or finer-grained ordering of the patients beyond the maximum time-to-event recorded in the train set. Another limitation is that the encoding for censored patients (see \textit{Section 2.1}) encourages the model to learn a decreasing survival probability over time beyond the censored time. While this is a natural and reasonable assumption in many cases, it is not always true. It is possible, for example, that a critically ill patient with early censoring may experience an event sooner.

\bibliographystyle{splncs04}
\bibliography{references}

\end{document}